# EXPLORING NON-AUTOREGRESSIVE END-TO-END NEURAL MODELING FOR ENGLISH MISPRONUNCIATION DETECTION AND DIAGNOSIS

*Hsin-Wei Wang*[1], *Bi-Cheng Yan*[1], *Hsuan-Sheng Chiu*[2], *Yung-Chang Hsu*[3], *Berlin Chen*[1]

[1] National Taiwan Normal University, Taiwan
[2] Chunghwa Telecom Laboratories, Taiwan
[3] EZAI, Taiwan
{hsinweiwang, bicheng, berlin}@ntnu.edu.tw; samhschiu@cht.com.tw; mic@ez-ai.com.tw

## ABSTRACT

End-to-end (E2E) neural modeling has emerged as one predominant school of thought to develop computer-assisted pronunciation training (CAPT) systems, showing competitive performance to conventional pronunciation-scoring based methods. However, current E2E neural methods for CAPT are faced with at least two pivotal challenges. On one hand, most of the E2E methods operate in an autoregressive manner with left-to-right beam search to dictate the pronunciations of an L2 learners. This however leads to very slow inference speed, which inevitably hinders their practical use. On the other hand, E2E neural methods are normally data-hungry and meanwhile an insufficient amount of nonnative training data would often reduce their efficacy on mispronunciation detection and diagnosis (MD&D). In response, we put forward a novel MD&D method that leverages non-autoregressive (NAR) E2E neural modeling to dramatically speed up the inference time while maintaining performance in line with the conventional E2E neural methods. In addition, we design and develop a pronunciation modeling network stacked on top of the NAR E2E models of our method to further boost the effectiveness of MD&D. Empirical experiments conducted on the L2-ARCTIC English dataset seems to validate the feasibility of our method, in comparison to some top-of-the-line E2E models and an iconic pronunciation-scoring based method built on a DNN-HMM acoustic model.

## 1. INTRODUCTION

The trend of globalization has underscored the importance of foreign language proficiency. This calls for continued development of effective computer-assisted pronunciation training (CAPT) systems to detect and provide feedback on mispronunciations of second-language (L2) learners, so as to improve their speaking proficiency through repeated practice. Meanwhile, more and more standardized examinations also adopt CAPT systems to assist proficiency evaluations on L2 speakers (e.g., TOEFL [1], AZELLA [2] and many others) for various use cases.

There has been a surge of research interest in design and development of effective methods to detect phone-level mispronunciation patterns of L2 learners by leveraging techniques originating from automatic speech recognition (ASR), which fall roughly into two categories. The first category of methods are pronunciation-scoring based ones [3-6]. Among them, goodness of pronunciation (GOP) and its descendants are the most celebrated instantiations [5, 7]. The principal idea behind GOP is to compute the ratio between the likelihoods of a canonical phone and the most likely pronounced phones predicted by an acoustic model via forced-alignment of the canonical phone sequence of a given text prompt to the speech signal uttered by a learner. A phone segment is identified as a mispronunciation if the corresponding likelihood ratio do not exceed a given threshold. GOP has been empirically shown to correlates well with human assessments.

The other category of methods makes the end-to-end (E2E) neural ASR paradigm straightforwardly applicable to mispronunciation detection. A de facto standard of these E2E neural methods is to first employ a free-phone recognition process [8-14] to dictate the possible sequence of phones produced by an L2 learner. To this end, an encoder-encoder attention model working in conjunction with an auxiliary connectionist temporal classification (CTC) objective [15-18] (denoted by CTC-ATT for short [19]) is frequently employed. Subsequently, the recognition transcript can be compared to the canonical phone sequence of the corresponding text prompt to simultaneously detect and give feedback on mispronunciations. Although current E2E neural methods built on various Transformer-based architectures [20] can offer good promise in comparison to conventional pronunciation-scoring based methods for CAPT, these methods are confronted with two intrinsic challenges. On one hand, most of Transformer E2E methods operate in an autoregressive manner with left-to-right beam search to dictate the pronunciations of an L2 learners. This inevitably leads to quite slow inference speed, making these methods struggle to satisfy the latency constraints for use in production applications. On the other hand, training E2E neural models is a data-hungry process. It would often result in performance degradation on mispronunciation detection and diagnosis (MD&D), when only a limited amount of nonnative training data is made available. Recognizing the above-mentioned issues, we present a novel MD&D method that capitalizes on non-autoregressive (NAR) E2E neural modeling with different acoustic feature extraction processes, which seeks to dramatically speed up the inference time while maintaining performance on par with the conventional Transformer-based E2E methods. Furthermore, we design and develop a pronunciation modeling network stacked on top of the NAR E2E models to further boost the efficacy of MD&D.

The rest of this paper is organized as follows. Section 2 sheds light on the fundamental framework and corresponding model components of our proposed method for MD&D. After that, the experimental setup and results are presented in Sections 3 and 4, respectively. Finally, we conclude this paper and envisage future research directions in Section 5.

## 2. PROPOSED METHOD

In this paper, we present a novel E2E neural method for mispronunciation detection and diagnosis, which consists of two ingredient elements, viz. a dictation model in tandem with a pronunciation model. The former manages to dictate the phone sequence of an L2 leaner's utterance, while the latter judges whether each dictated phone is a correct pronunciation or not given its confidence score and the prompt text corresponding to the utterance. Figure 1 shows a schematic illustration of our proposed method for mispronunciation detection and diagnosis.

### 2.1. Dictation model (DM)

Transformer-based E2E methods stemming from ASR, such as CTC-ATT, have been applied with good success to mispronunciation detection and diagnosis. Their operations however proceed in an autoregressive manner to dictate the phone sequence corresponding to an L2 leaner's utterance. More specifically, the decoder of these methods employs left-to-right progressive beam search to predict the next probable phone by conditioning on the topmost historical phone sequences decoded so far and the high-level acoustic context attained from the encoder output. The inference time for these Transformer-based E2E methods is usually slow (viz. scaling quadratically with the length of the output phone sequence), which would make them unappealing for real use cases. On a separate front, there is an emerging area of active research dedicated to develop non-autoregressive (NAR) E2E modeling to speed up the inference time of E2E neural machine translation (MT) and ASR methods, such as Mask-CTC [21], LASO [22] and Align-Refine [23], to name just a few. However, as far as we are aware, CAPT tasks with NAR E2E modeling remain relatively underexplored. In this regard, we explore in this paper a novel use of Mask-CTC to build the dictation model in replace of CTC-ATT.

Succinctly stated, the Mask-CTC dictation model that consists of three ingredient components (*cf.* Figure 1) is trained with a joint CTC and mask-predict objective [21]. In the inference stage, the dictated phone sequence of an L2 learner is initialized with the CTC's output phone sequence: A Conformer-based encoder component [24] first converts frame-wise acoustic feature vectors into intermediate frame-wise phonetic representations, followed by frame-wise latent alignment between intermediate frame-wise phonetic representations and the output phone sequence. The dictated phones with low confidence scores are masked subsequently. In turn, each of these masked phones is respectively predicted by conditioning on other high-confidence phones in its both-side context using a Transformer-based conditional masked language model (CMLM) [25]. As such, the Mask-CTC dictation model can not only take advantage of the parallel computation ability of Conformer (viz. its Transformer part) to achieve fast dictation, but also leverage the synergistic power of CTC and CMLM to reach dictation performance competitive to CTC-ATT.

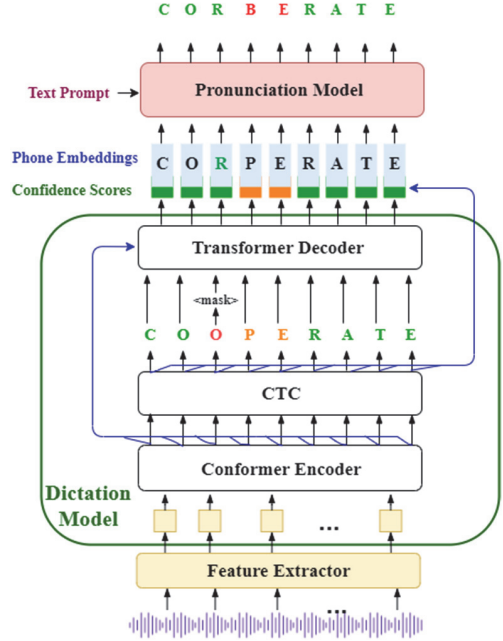

*Figure 1: A schematic illustration of our proposed method for mispronunciation detection and diagnosis.*

### 2.2. Pronunciation model (PM)

It would be tempting to simply compare the output phone sequence generated by the dictation model to the canonical phone sequence of the corresponding text prompt to readily detect and give feedback on mispronunciations. However, as is well known, L2 English learners may have a wide variety of accents, which is usually influenced by their mother-tongue languages. In addition, due to the lack of sufficient labeled speech data of L2 speakers for model estimation, a single, one-size-fits-all dictation model (or ASR model) is prone to overfitting and thus would provide an imperfect MD&D decision. In view of the above observations, we seek to defer the decision to an extra pronunciation model (*cf.* Figure 2), which is anticipated to better judge whether each dictated phone is a correct pronunciation or not, given its confidence score and the prompt text corresponding to the utterance. To flesh out this idea, for each L2 learner's utterance we first convert the phone sequence generated by the Mask-CTC dictation model into a phone embedding sequence. Each phone embedding of the sequence is concatenated with its respective confidence score. The augmented phone embedding sequence in turn is fed into the Transformer-based encoder part to obtain an intermediate phone embedding sequence. Meanwhile, a single- or multiple-layer GRU (short for Gated Recurrent Unit) neural network takes input the canonical phone embedding sequence of the text prompt to produce an utterance-level embedding vector of the text prompt, viz. text prompt embedding. After that, we devise a text prompt-modulated gating (PMG) mechanism to perform fine-grained prompt-aware modulation of the intermediate phone embedding, with the hope to further enhance the discriminating capability of the pronunciation model. The formulations of PMG are concisely expressed as follows:

$$\mathbf{g}_s = \text{Sigmoid}([\mathbf{h}_S; \mathbf{a}_E]W_1 + \mathbf{b}_1) \quad (1)$$

$$\mathbf{h}'_S = \text{Relu}(\mathbf{h}_S + [\mathbf{h}_S \odot \mathbf{g}_s]W_2 + \mathbf{b}_2) \quad (2)$$

where $\odot$ represents the elementwise dot production and $W_1$, $W_2$, $\mathbf{b}_1$ and $\mathbf{b}_2$ are trainable parameters. The functionality of Eq. (1) is to model the similarity and discrepancy between an intermediate phone embedding $\mathbf{h}_S$ at timestep $s$ and the utterance-level prompt embedding $\mathbf{a}_E$, thereby producing a gating vector $\mathbf{g}_s$ with the Sigmoid function. Furthermore, Eq. (2) seeks to obtain a modulated intermediate phone embedding $\mathbf{h}'_S$ with the gating vector $\mathbf{g}_s$ and the Relu function in succession. Finally, an additional single- or multi-layer feedforward neural network (FFN) with the Softmax normalization, stacked on top of the PMG component and taking the modulated intermediate phone embedding sequence as the input, is employed to make the ultimate MD&D decision.

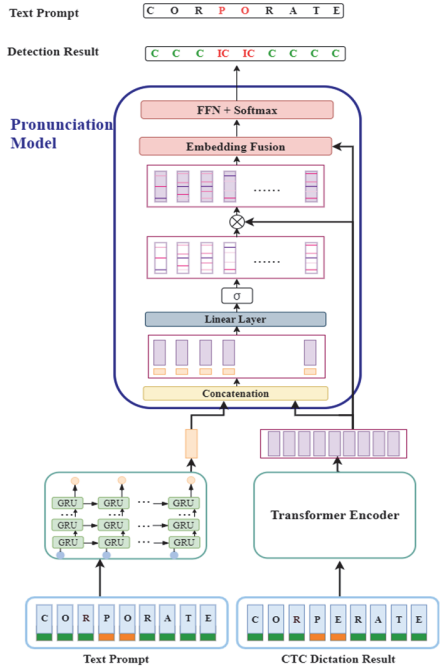

*Figure 2: A schematic illustration of the pronunciation model.*

Table 1. Statistics of the experimental speech corpora.

|  |  | Spks. | Utts. | Hrs. |
|---|---|---|---|---|
| TIMIT | Train | 462 | 3,696 | 3.15 |
|  | Dev | 50 | 400 | 0.34 |
|  | Test | 24 | 192 | 0.16 |
| L2-ARCTIC | Train | 17 | 2,549 | 2.66 |
|  | Dev | 1 | 150 | 0.12 |
|  | Test | 6 | 900 | 0.88 |

Table 2. The phone error rate (%) and RTF results achieved by our method (merely with the dictation model) and other E2E MD&D methods.

|  | L2-ARCTIC | |
|---|---|---|
|  | PER | RTF |
| CTC-ATT (beam size=1) | 15.2 | 3.10 |
| CTC-ATT (beam size=10) | 14.9 | 13.85 |
| Mask-CTC (FBANK) | 22.6 | 0.05 |
| Mask-CTC (W2V) | 15.9 | 0.05 |

## 3. EXPERIMENTAL SETUP

We present in this section the experimental datasets and the metrics we employed to evaluate various MD&D methods for mispronunciation detection and diagnosis.

### 3.1. Experimental Corpora

A series of mispronunciation detection experiments were conducted the L2-ARCTIC benchmark corpus [26]. L2-ARCTIC is an open-access L2-English speech corpus compiled for research on CAPT, accent conversion, and others. About 3,600 utterances of 24 non-native speakers (12 males and 12 females; equipped with manual transcripts) across different nationalities, some of which contained mispronounces, collectively constituted L2-ARCTIC (Version 4). These speakers were made up of six mother-tongue languages, including Hindi, Korean, Mandarin, Spanish, Arabic and Vietnamese. In addition, a limited amount of native (L1) English speech data compiled from the TIMIT corpus [27] (composed of 630 speakers) was pooled together with the training set of L2-ARCTIC for the estimation of the various E2E MD&D models and the DNN-HMM acoustic model of the GOP-based method. To follow the setup of TIMIT, at training time, these models all employed an inventory of 48 distinct canonical phones, each of which was supplemented with a peculiar "anti-phone" for the purpose of labeling its corresponding mispronounced segments in L2-ARCTIC that were caused by non-categorical or distortion errors, viz. approximating L2 phones with L1 (first-language) phones, or erroneous pronunciation patterns in between. At test time, the original canonical phone inventory was consolidated into a new inventory of 39 distinct canonical phones, so as to make our experimental setup in line with previous work [11-12]. Furthermore, the setting of the mispronunciation detection experiments on L2-ARCTIC followed the recipe provided by [12]. Summary statistics of the TIMIT and L2-ARCTIC datasets are shown in Table 1.

The input to the dictation model was composed of 80-dimensional Mel-filter-bank feature vectors (denoted by FBANK for short) extracted from frames of 25 msec length and 10 msec shift. In addition, we also employed wav2vec 2.0 (a pretrained neural network) [28] as an alternative front-end feature extractor (denoted by W2V for short), which has achieved a good level of success for improving the performance of numerous ASR tasks, especially when available training data is of limited size.

Table 3. Performance evaluations on correct pronunciation and mispronunciation detection.

| | Correct Pronunciation Detection (CD) | | | Mispronunciation Detection (MD) | | |
|---|---|---|---|---|---|---|
| | PR | RE | F1 | PR | RE | F1 |
| GOP | 91.97 | 90.98 | 91.47 | 46.99 | 50.15 | 48.52 |
| CTC-ATT | 91.04 | 92.24 | 91.73 | 47.55 | 42.94 | 45.13 |
| CNN-RNN-CTC | 93.88 | 79.97 | 86.37 | 34.88 | 67.29 | 45.94 |
| Mask-CTC (w/o PM) | 91.62 | 90.94 | 91.28 | 45.73 | 47.87 | 46.77 |
| Mask-CTC (w/ PM) | 91.70 | 90.80 | 91.25 | 45.66 | 48.46 | 47.02 |

### 3.2 Performance evaluation metrics

The default evaluation metric employed in this paper for correct pronunciation and mispronunciation detection is the F1-score (F1), which is a harmonic mean of precision (PR) and recall (RE). Furthermore, the diagnosis accuracy rate (DAR) is calculated by the ratio between the number of correct diagnosis (CD) and the sum of correct diagnosis (CD) and incorrect diagnosis (ID). More detailed information about our experimental setup is made available at https://github.com-/HsinWeiWangYH/PMG.

## 3. EXPERIMENTAL RESULTS

At the outset, we report on the phone error rate (PER) and real-time factor (RTF) performance of our method (with the dictation model alone), as well the conventional CTC-ATT (with different beam sizes) method, on the L2-ARCTIC test set. The corresponding results as shown in Table 2, where the RTF results were measured by using a normal computer equipped with Intel(R) Core (TM) i9-9900KF CPU @ 3.60GHz. Here we exclude the time consumption of feature extraction for the RTF calculation. Furthermore, CTC-ATT employed a Conformer-based encoder and the FBANK-based acoustic features. From Table 2, we can make the following two observations. First, CTC-ATT (with the beam size equal to 1 or 10) delivers considerably better performance than Mask CTC with the FBANK-based acoustic features (viz. Mask-CTC (FBANK)), at the cost of relatively high RTFs in relation to the latter. Second, Mask-CTC with the W2V-based acoustic features (viz. Mask-CTC (W2V)) bridges the PER performance gap dramatically to CTC-ATT (beam size=10), and simultaneously is 277 times faster than the latter in terms of RTF. In the following experiments, Mask-CTC (W2V) will be taken as our default dictation model, unless otherwise stated.

In the second set of experiments, we turn to evaluating the efficacy of two different instantiations of our method for correct pronunciation and mispronunciation detection, in comparison to CTC-ATT, the so-called CNN-RNN-CTC method [10] and the celebrated GOP-based method. The corresponding results are shown in Table 3, from which we can make some interesting observations. First, Mask-CTC using the dictation model alone (viz. Mask-CTC (w/o PM)) not only performs on par with CTC-ATT, but also significantly reduces the processing time required for correct pronunciation and mispronunciation detection (*cf.* Table 2). Second, Mask-CTC (w/ PM), viz. using both the dictation model and pronunciation model, can lead to a better F1-score result than Mask-CTC (w/o PM) on mispronunciation detection, which comes at the cost of a very slight degradation of the F1-score result on correct pronunciation detection. Note here that the processing time of the pronunciation model is almost negligible in terms of RTF. This also confirms our conjecture that our NAR E2E method can be a promising alternative to conventional CTC-ATT based methods that generally execute in an autoregressive manner. Third, the performance levels of our method and CTC-ATT are still slightly inferior to GOP, which is due probably to the fact that the amount of non-native training data used in our experiments is small and the numbers of parameters of our method and CTC-ATT are much larger than that of GOP. As a final note, our method and CTC-ATT can readily provide diagnosis on the mispronounced phone segments of an L2 learner's utterance, with diagnosis accuracy rates of 67.19% and 73.22% for mispronounced utterances, respectively.

## 4. CONCLUSION AND FUTURE WORK

In this paper, we have presented a novel use of non-autoregressive (NAR) E2E modeling framework for mispronunciation detection and diagnosis (MD&M), showing considerable promise in relation the conventional CTC-ATT and GOP methods. Our method also can dramatically speed up the inference time, which indeed confirms the practical feasibility the NAR E2E approach to various potential CAPT tasks. As to future work, we would like to explore more fine-grained acoustic, prosodic, accent, or suprasegmental pronunciation phenomena for NAR E2E approaches to mispronunciation detection and diagnosis [29, 30].

## 6. ACKNOWLEDGEMENTS

This research is supported in part by Chunghwa Telecom Laboratories under Grant Number TL-110-D301, and by the National Science Council, Taiwan under Grant Number MOST 109-2634-F-008-006- through Pervasive Artificial Intelligence Research (PAIR) Labs, Taiwan, and Grant Numbers MOST 108-2221-E-003-005-MY3 and MOST 109-2221-E-003-020-MY3. Any findings and implications in the paper do not necessarily reflect those of the sponsors.